\def\year{2019}\relax
\documentclass[letterpaper]{article} 
\usepackage{aaai19}  
\usepackage{times}  
\usepackage{helvet}  
\usepackage{courier}  
\usepackage{url}  
\usepackage{graphicx}  
\usepackage{tabularx,booktabs}
\newcolumntype{C}{>{\centering\arraybackslash}X} 
\newcolumntype{s}{>{\hsize=.1\hsize}X}
\usepackage{amsmath,amssymb}
\DeclareMathOperator{\E}{\mathbb{E}}
\begin{document}
%
\title{Hierarchical Deep Feature Learning For Decoding Imagined Speech From EEG}
\author{Pramit Saha and Sidney Fels\\
Department of Electrical and Computer Engineering, University of British Columbia, Vancouver, Canada\\
pramit@ece.ubc.ca, ssfels@ece.ubc.ca
}
\maketitle
\begin{abstract}
We propose a mixed deep neural network strategy, incorporating parallel combination of Convolutional (CNN) and Recurrent Neural Networks (RNN), cascaded with deep autoencoders and fully connected layers towards automatic identification of imagined speech from EEG. Instead of utilizing raw EEG channel data, we compute the joint variability of the channels in the form of a covariance matrix that provide spatio-temporal representations of EEG. The networks are trained hierarchically and the extracted features are passed onto the next network hierarchy until the final classification. Using a publicly available EEG based speech imagery database we demonstrate around 23.45\%  improvement of accuracy over the baseline method. Our approach demonstrates the promise of a mixed DNN approach for complex spatial-temporal classification problems.
\end{abstract}

\section{Introduction}
In the past decade, numerous methods have been proposed to decode speech and motor-related information from electroencephalography (EEG) signals for Brain-Computer Interface (BCI) applications. However, EEG signals are highly session-specific; infested with noises and artifacts. Interpreting active thoughts underlying vocal communication involving labial, lingual, naso-pharyngeal and jaw motion, is even more challenging than inferring motor imagery, since utterances involve higher degrees of freedom and additional functions in comparison to hand movement and gestures. As a result, it is extremely challenging to recognize phonemes, vowels and words from single-trial EEG data. The reported classification accuracy of existing methods~\cite{nguyen2017inferring,min2016vowel,dasalla2009single} are not satisfactory, showing that manual handcrafting of features or traditional signal processing algorithms lack sufficient discriminative power to extract relevant features for classification.

This work addresses these issues by implementing a deep learning based feature extraction scheme, targeting classification of speech imagery EEG data corresponding to vowels, short and long words. It is mentionworthy that most of the previous works applied to vowels and phonemes classification show a degradation of performance when applied to words and vice versa. Therefore, this is the first work that aims to automatically learn a discriminative EEG manifold applicable to both word and vowel/phoneme based classification at the same time, using deep learning techniques. 

\section{Proposed framework}
The problem of categorizing EEG data based on speech imagery can be formulated as a non-linear mapping~ $\hat{\textit{\textbf{f}}}$ of a multivariate time-series input sequence ${\textbf{X}^{c}_t}$ to fixed output $\textbf{\textit{y}}$, \textit{i.e}, mathematically ~$\hat{\textbf{\textit{f}}}$~:~ $\textbf{X}^{c}_t~ \longrightarrow~\textbf{\textit{y}}$. 

We found that well-known deep learning techniques, like fully connected networks, CNN, RNN, autoencoders \textit{etc.} fail to individually learn such complex feature representations from single-trial EEG data. Also, our investigation demonstrated that it is crucial to capture the information transfer between the electrodes rather than using the multi-channel high-dimensional EEG data that otherwise needs large training times and resource requirements. 
Therefore, instead of utilizing raw EEG, we compute channel cross-covariance, resulting in positive, semi-definite matrices encoding the joint variability of electrodes. We define channel cross-covariance (CCV) between any two electrodes $c_1$ and $c_2$ as:
$Cov({X}^{c_{1}}_t,{X}^{c_{2}}_{t+\tau})=\displaystyle \E[X^{c_{1}}(t)-\mu_{X^{c_{1}}}(t)][X^{c_{2}}(t+\tau)-\mu_{X^{c_{2}}}(t+\tau)]$.

This is particularly important because higher cognitive processes underlying speech synthesis and utterances involve frequent information exchange between different parts of the brain. Hence, such matrices often contain more discriminative features and hidden information than mere raw signals. 
\begin{figure}
\includegraphics[width=18.6cm,height=6.1cm,keepaspectratio]{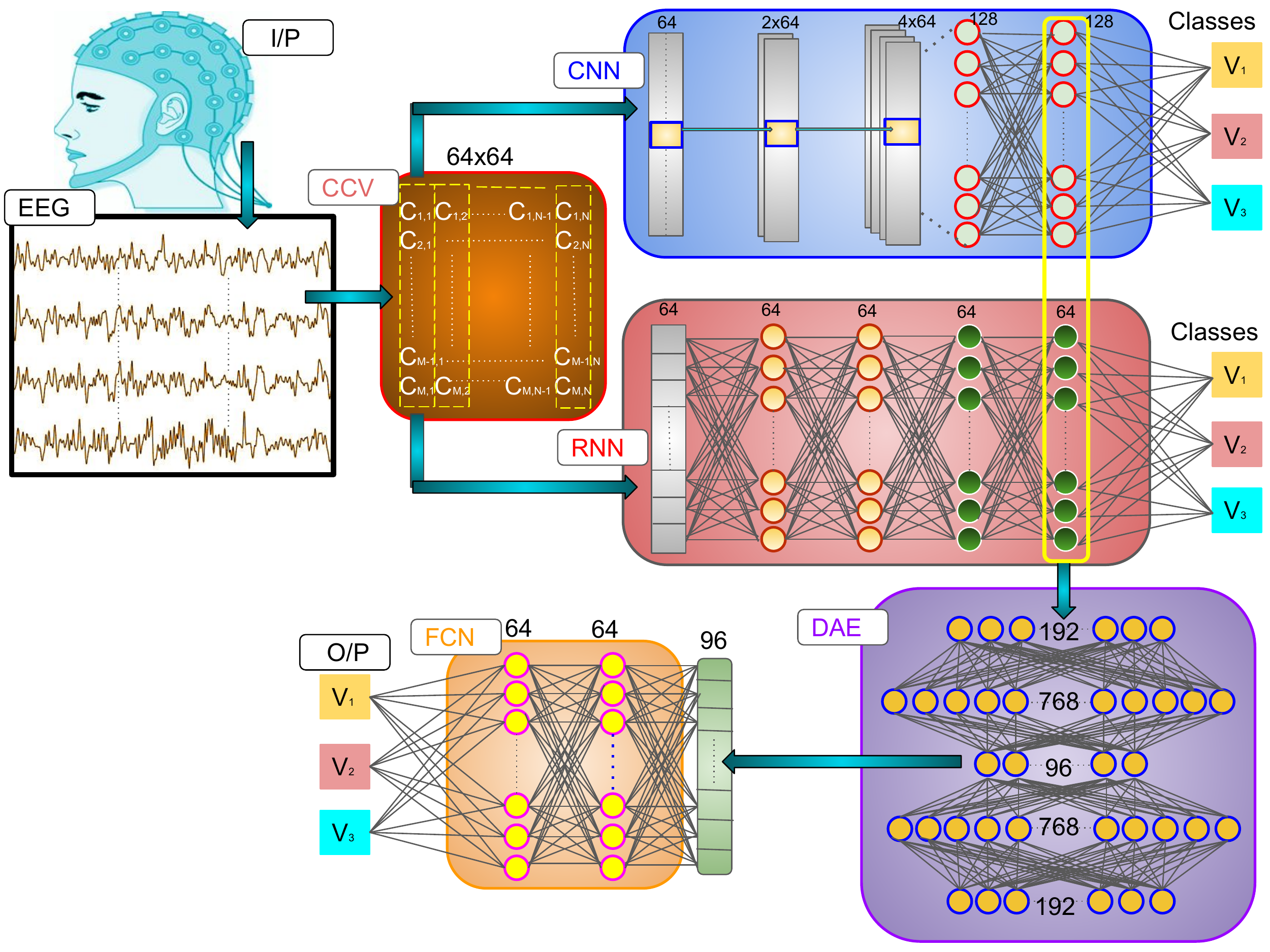}
\caption{Overview of the proposed approach}
\end{figure}
Besides, cognitive learning process underlying articulatory speech production involves incorporation of intermediate feedback loops and utilization of past information stored in the form of memory as well as hierarchical combination of several feature extractors. To this end, we develop our mixed neural network architecture composed of three supervised and single unsupervised learning step (Fig 1).

In order to decode spatial connections between the electrodes from the channel covariance matrix, we use six-layered 1D convolutional networks stacking two convolutional and two fully connected hidden layers with ReLU as activations. The $k^{th}$ feature map at a given CNN layer with input $x$, weight matrix $W^{k}$ and bias $b_k$ is obtained as: $h^{k}=ReLU(W^{k}*x+b_k)$. The network is trained with the corresponding labels as target outputs, optimizing cross-entropy cost function via AdamOptimizer. 

In parallel, we apply a six-layered recurrent neural network on the channel covariance matrices to explore the hidden temporal features of the electrodes. It consists of two fully connected hidden layers, stacked with two LSTM layers and is trained in a similar manner as CNN. 

Since these parallel networks are trained individually and $5^{th}$ layer of both the networks has a direct relationship with respective output layers, we claim that these layers are powerful discriminative spatial and temporal representations of the data. Therefore, we concatenate these feature vectors to form joint spatio-temporal encodings of covariance matrix. 

The second level of hierarchy encompasses unsupervised training of deep autoencoders (DAE) having two encoder-decoder layers, with mean squared error (MSE) as the cost function. This leads to further dimensionality reduction of the spatio-temporal encodings \cite{zhang2018converting}. 

At the third level of hierarchy, the discrete latent vector representation of the deep autoencoder is fed into a two-layered fully connected network (FCN) followed by softmax classification layer. This is again trained in a supervised manner similarly as the CNN network, to output the final predicted classes corresponding to the speech imagery. 

\section{Experiments and Results}
We evaluate our model on the publicly available imagined speech EEG dataset ~\cite{nguyen2017inferring}. It consists of imagined speech data corresponding to vowels, short words and long words, for 15 healthy subjects. The short words included in the study were `\textit{in}', `\textit{out}' and `\textit{up}'; the long words were `\textit{cooperate}' and `\textit{independent}'; the vowels were \textit{/a/}, \textit{/i/} and \textit{/u/}. 
The participants were instructed to pronounce these sounds internally in their minds, avoiding muscle movements and overt vocalization. 

\begin{figure}
\includegraphics[width=8.5cm,height=8.5cm,keepaspectratio]{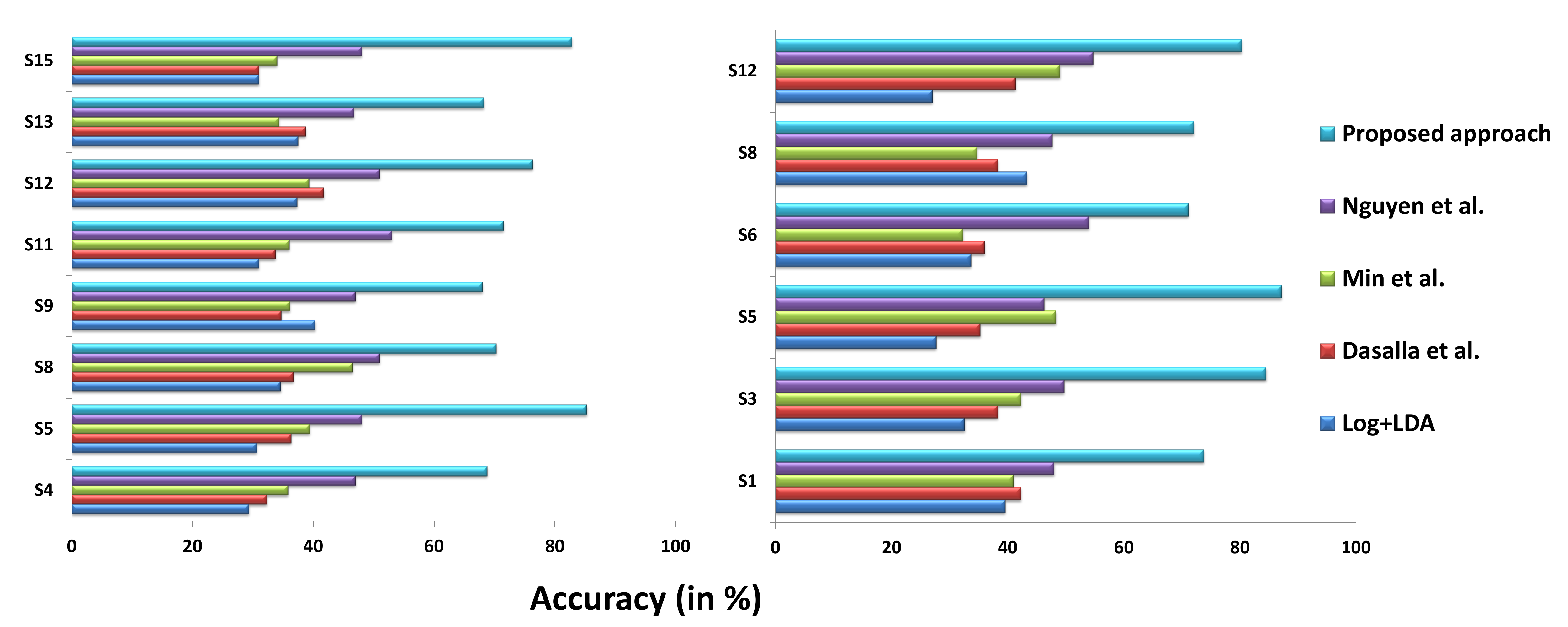}
\caption{Performance comparison for all subjects on vowels (left) and short words (right)}
\end{figure}

We trained the networks with 80\% of the data in the training set and the remaining 20\% in the validation set. Table 1 shows the average test accuracy for all subjects corresponding to long word classification. Fig 2 provides a comparison of test accuracy of our approach with others on vowels and short word classification.
Our results illustrate that our model achieves significant improvement over the other methods, thereby validating that deep learning based hierarchical feature extraction can learn a better discriminative EEG manifold for decoding speech imagery. 
\begin{table}[t]
  \caption{Classification accuracy on long words}
  \label{tab:word_styles}
  \centering
  \begin{tabular}{llllllll}
    \toprule
    Method&S2&S3&S6&S7&S9&S11\\
    \toprule 
    Nguyen et al.& 70.0 & 64.3&	72.0&	64.5&	67.8&	~58.5             \\ 
    Proposed &77.5& 90.7&	73.7&	86.8&	80.1&	~71.1\\
        \midrule
\end{tabular}
\end{table}
\section{Conclusion}
Towards recognizing active thoughts from EEG corresponding to vowels, short words and long words, this work presents a novel mixed neural network strategy as a combination of convolutional, recurrent and fully connected neural networks stacked with deep autoencoders. The network is trained hierarchically on a channel covariance matrix for categorizing respective EEG signals to the  imagined speech classes. Our model achieves satisfactory performance with different types of target classification, on different subjects and hence can be considered as a reliable and consistent approach for classifying EEG based speech imagery. 
\bibliographystyle{aaai}
\bibliography{Saha_AAAI.bbl}

\end{document}